# A New Weakly Supervised Learning Approach for Real-time Iron Ore Feed Load Estimation

Li Guo[1], Yonghong Peng[1], Rui Qin[1], Bingyu Liu[2]

[1]Department of Computing and Mathematics, Manchester Metropolitan University, United Kingdom

Email: L.Guo@mmu.ac.uk, Y.Peng@mmu.ac.uk, R.Qin@mmu.ac.uk

[2]Ansteel Mining Corporation, Anshan, Liaoning, P.R China

Email: liubingyu@ansteel.com.cn

**Abstract**— Iron ore feed load control is one of the most critical settings in a mineral grinding process, directly impacting the quality of final products. The setting of the feed load is mainly determined by the characteristics of the ore pellets. However, the characterisation of ore is challenging to acquire in many production environments, leading to poor feed load settings and inefficient production processes. This paper presents our work using deep learning models for direct ore feed load estimation from ore pellet images. To address the challenges caused by the large size of a full ore pellets image and the shortage of accurately annotated data, we treat the whole modelling process as a weakly supervised learning problem. A two-stage model training algorithm and two neural network architectures are proposed. The experiment results show competitive model performance, and the trained models can be used for real-time feed load estimation for grind process optimisation.

## I. INTRODUCTION

The ball mill grinding circuit (GC) plays an essential role in production in the mineral industry. The low-grade iron ores must go through a beneficiation process for further concentration (Figure 1).

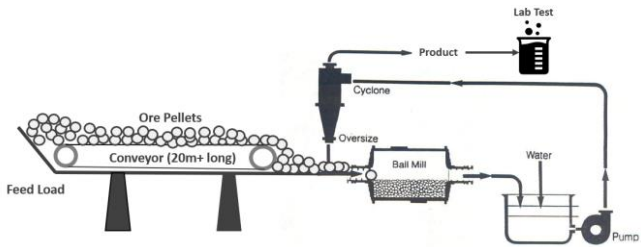

*Figure 1: The Single State Ball Mill Grinding Circuit Process*

The quality of the product depends on the control of a range of setpoints within a GC process, such as ore load (feed load), mill pressure, water level, and pump pressure, as shown in Table 1.

| Symbol | Setpoints |
|---|---|
| $s_1$ | Feed load (range from 120 tons to 146 tons). This setpoint controls how much ore pellets are fed onto the conveyor |
| $s_2$ | Ball mill water level. Water is regularly injected into the ball mill to ensure the grinding operation does not damage the ball mill. |
| $s_3$ | Hydraulic inlet pressure. When the hydraulic pressure of the hydro-cyclone is lower than the stable value, the fine slurry cannot be separated, and it causes a pulp overflow. |
| $s_4$ | Dilution water level setpoint for the pump pool. |
| $s_5$ | Slurry density setpoint |

*Table 1: Setpoints for a Grinding Circuit*

In mineral industries in China, the GC control practices are manually designed based on the laboratory analysis result of the product as feedbacks for adjusting those setpoints. Amongst all possible setpoints, the feed load setpoint needs to be modified as the others' adjustments require a highly knowledgeable and experienced human operator. Also, frequent change of those setpoints considerably reduces the lifetime and stability of the equipment. In other words, the product quality largely depends on the ore characteristics and the feed load settings. For example, for ore pellets that are hard to grind, the feed load is set low to ensure the product quality, and that is set higher for easier-grinding ore for better product throughput.

Even with this simplified mechanism (single setpoint feedback control), the current practice often leads to volatile product quality or inefficient productivity due to the considerable latency between the actual control operation and the lab result generation. The laboratory analysis is usually conducted every four hours because of the resources required and the chemical process involved. This interval is much longer than the time needed for a single GC-cycle (20-30 minutes, from ore feed to final product). Before the latest lab test result arrives, human operators must use their experience to set feed load for a better-quality product. However, as a more typical case, inexperienced human operators often use the latest lab test result as the only indicator for feed load adjustment for the upcoming ore pellets. It is not surprising to see that such practices lead to poor production performance. Therefore, there is a need for methods that either provide instant feedback on product qualities or provide real-time guidance on tuning the feed load to ensure acceptable product quality.

The mainstream methods/equipment deployed are based on ray sensors such as XRT and XRF[1]–[3] to the ore's characteristics to provide real-time information. Although ray-sensors come with high classification accuracies, their cost and radiation are very high. In recent years, as alternatives, vision/image-based approaches have attracted significant interest from both academia and industry because of their low cost, easy/flexible installation and maintenance process, and competitive performance compared to the other solutions. Many attempts have been made to apply machine learning or deep learning-based[4] methods for ore image analysis. These early works primarily focus on the tasks of ore sorting[5]–[7], ore grade estimation [8]–[11] and particle/blast fragment size estimation[10], [12]–[15]. Despite the promising results, there exist significant differences between the GC process and the other ore production processes. For instance, for ore sorting based analysis, the size of ore rock is usually bigger. Also, in the early reported work, ore rocks are often arranged on a conveyor or testbench sparsely with no overlap for image capturing. Images are then fed into a machine learning model for extracting remarkable features of each rock. In addition, ore images are all manually annotated to ensure the best-quality training samples are used. The above conditions, especially the last one, barely hold in practice in a grinding process. There are a few GC specific challenges that cannot be easily addressed with the existing methods.

First, in a grinding process, the ore is crushed and stir-mixed as an initial step. The ore pellets produced should have no significant difference in size and are much smaller than that from the ore sorting stage. They are then transited to the ball mill through the conveyor in heaps with random occlusion, as shown in Figure 2 (top row). This makes it hard for a simple model to extract individual pellet features or directly predict ore characteristics. In other work, microscopy images are often used for ore particles with small sizes instead of from commodity cameras. Second, apart from the lab test result, there is no ground truth of the ore characteristics (ore hardness, ore types) available in the GC production environment. The ore type data from the upstream stock are often inaccurate or wrong due to many uncontrolled operations. Third, as only a single lab test result is generated every 4 hours, it is costly and time-consuming to collect enough accurately annotated training samples, covering various types of ore pellets under different conditions, such as lighting (Figure 2, bottom row) feed load. Furthermore, knowing ore characteristics does not optimise the grinding process directly, as such information must be transferred into executable operations through additional modelling.

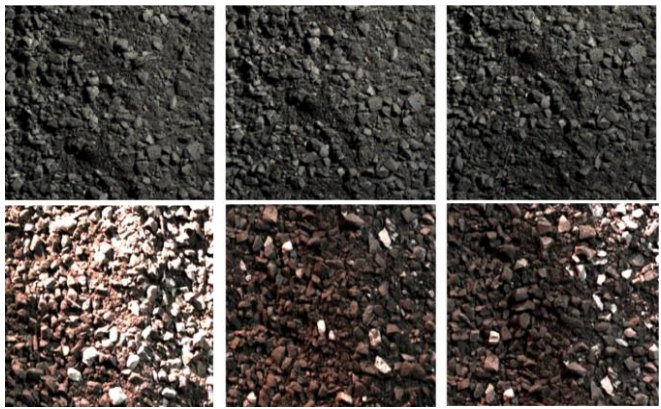

*Figure 2: Example Images of Ore Pellets on the Conveyor. The top row shows the ore pellets under normal lighting conditions, while the bottom row shows the ore pellets under different lighting conditions.*

To address the above challenges, in this paper, we present an image-based approach for real-time ore feed load estimation using deep learning models and weakly supervised learning methods. Our work's fundamental hypothesis is that it is possible to directly infer the optimal feed load settings from ore pellet images without explicitly modelling the ore characteristics (latent variables).

The main contributions of this work are as follows.

1> To the best of our knowledge, this work is the first attempt that applies image-based methods for direct ore load estimation in the ball mill grinding process. Due to its low cost and easy deployment/maintenance, it has clear potential for improving the productivity of a grinding process, thus having direct economic benefits for mineral industries.
2> We propose to use a new two-stage modelling algorithm, with which two neural network architectures are designed for both patch-image level feature extraction and global classification.
3> We propose treating the problem as a weakly supervised learning problem and demonstrating how smaller patch images with wrong labels can be utilised directly for model training with theoretical and empirical studies.
4> We conduct a series of experiments and provide empirical evidence, showing that the proposed work not only perform well with the testing data but also yields a significant economic boost in actual production.

The remainder of this paper is organised as follows. Section II explains our methodology, including problem formulation, the main algorithm, the network architectures and their design rationales. Experiments and results analysis are presented in section III. Section IV discusses challenges and restrictions for evaluation and demonstrates how this work is evaluated in a real production environment. Conclusion and future work are discussed in section V.

## II. METHODOLOGY

### A. Problem Analysis and Formulation

In the literature, the optimal control of a GC process is usually modelled as a dynamic system either through direct dynamic system modelling [6]–[8] or data-driven models [9]–[11], since many setpoints determine the production performance together. For this work, as explained earlier, only the feed load setpoint is considered for production performance (as required by the production process). Hence the whole GC process can be formulated as a probability distribution: $P(Q_{t+n}|s_t, O_t)$, where $s_t$ is the feed load setting at the time $t$, $O_t$ is a set of latent variables that denote the ore pellets' characteristics (type, hardness, sizes) at time $t$, and $P(Q_{t+n}|s_t, O_t)$ denotes the probability measure of final product quality values at a later time $t+n$, given $s_t$ and $O_t$. In addition, a GC process typically has a product quality threshold ($q_m$), which governs the whole process. It is the most important constraint of the system that has substantial economic impacts. Other production performance indicators, such as product throughput, get measured only when the $q_m$ constraint is satisfied. Putting everything together, we can write the objective as:

$$argmax_{s_t}(s_t * P(Q_{t+n} = q_m|s_t, O_t)) \quad (1)$$

Equation (1) can be rewritten to:

$$argmax_{s_t}\left(s_t * \frac{P(s_t, O_t|Q_{t+n} = q_m)*P(Q_{t+n}=q_m)}{p(s_t,O_t)}\right) \quad (2)$$

As $q_m$ is the constraint that must be satisfied, to maximise the equation (2), we could make $P(Q_{t+n} = q_m) = 1$. The denominator, $P(s_t, O_t)$ can also be ignored as it is for the normalisation purpose. Therefore, equation (2) is simplified to:

$$argmax_{s_t}(s_t * P(s_t, O_t|Q_{t+n} = q_m)) \quad (3)$$

We still need to work out the value of $O_t$ and this is where the ore pellet images come into play. The assumption made here is that $O_t$ can be approximated by a function $w(I_t) \approx O_t$, where $I_t$ denotes the image captured at the time $t$. After substituting $O_t$ with $w(I_t)$, we drive the final objective function:

$$argmax_{s_t}(s_t * P(s_t, w(I_t)|Q_{t+n} = q_m)) \quad (4)$$

Maximising equation (4) is equivalent to maximising $P(s_t, w(I_t)|Q_{t+n} = q_m)$, as after being normalised, the variance of the component $s_t$ is small and has little impact on the result of the whole equation.

It is not difficult to imagine using a neural network model for estimating this distribution. However, the key challenge for implementing $P(s_t, w(I_t)|Q_{t+n} = q_m)$ arises from the difficulties in collecting enough and appropriate ($s_t, I_t$) samples that make statistical sense. We have briefly discussed this issue earlier. $Q_{t+n}$ is generated every 4 hours, which means, maximumly, only six image samples can be collected. It should be noted that even for these six images, their corresponding $Q_{t+n}$ values might be smaller than the $q_m$ threshold and therefore cannot be used for solving equation (4).

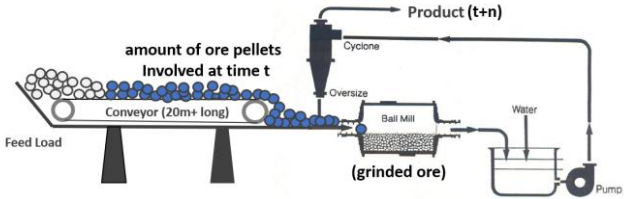

*Figure 3: The Amounts of Ore Pellets Required for a Single Product Result*

Furthermore, another more challenging issue is from the potential size of an image sample ($I_t$). The product generated at $t + n$ is ground from a cluster of ore pellets, which requires a huge image receptive field (much more than 20 meters long) to cover (shown in Figure 3). It is practically infeasible to use a single image as a representation for the involved ore samples. Sequence analysis of a video clip neither suits this application, as there is no actual temporal relation between the ore pellets from each video frame.

To address the above two issues, we propose to use a set of smaller images $\mathcal{P}_t = [\wp_i|\wp_{t-m}, ..., \wp_{t+m}]$ to replace $I_t$. The $\mathcal{P}_t$ can be understood as subsamples of $I_t$ [1](a set of small patches from a large image). In this way, we can also collect substantially more images (2m times more per day) for later training. After substituting $I_t$ with $\mathcal{P}_t$, the $w(I_t)$ becomes $w(I_t) = w_1(w_p(\wp_i))$, $s_t$ becomes $s_t \approx \frac{\sum_{i=t-m}^{i=t+m} s_i}{2m}$ and the equation (4) becomes:

$$argmax_{s_t}\left(s_t * P(s_t, w_1(w_p(\wp_i))|Q_{t+n} = q_m)\right) \quad (5)$$

which can then be transformed to:

$$argmax_{s_t} * \omega(P(s_{1t}, w_p(\wp_i)|Q_{t+n} = q_m)) \quad (6)$$

We can see, in equation (6), $\omega(P(s_t, w_p(\wp_i)|Q_{t+n} = q_m)$ is used to replace the $P(s_t, w_1(w_p(\wp_i))|Q_{t+n} = q_m)$ in equation (5). We believe there exists a function $\omega$, which approximates the probability of $P(s_t, w_1(w_p(\wp_i))|Q_{t+n} = q_m)$ using a list of joint probabilities of $P(s_t, w_p(\wp_i)|Q_{t+n} = q_m)$.

At last, since we are unable to find the value of $P(s_t, w_p(\wp_i)|Q_{t+n} = q_m)$ for each $\wp_i$, we first assume they share the same probability with $I_t$ as a starting point.

$$P(s_t, w_p(\wp_i)|Q_{t+n} = q_m) = P(S_t, I_t|Q_{t+n} = q_m) \text{ (asm. 1)}$$

This assumption gets refined in the next step of our work with weakly supervised learning methods.

### B. Training Models with Weakly-Supervised Learning

To solve the function $\omega$, we could use a convolutional neural network (CNN) [22] model. In the ideal situation, to train such a model, we can collect enough small images patches ($\wp_i$) and label each of them with a $s_t$. All unique $s_t$ values (generally under 20 in the actual production line) can be considered as classes that the model tries to classify. A SoftMax function is then used to generate a probability for each $s_t$ class, and we only need to choose the one with the maximum value. However, because of the assumption (asm. I) made, the training sample ($p_i, s_t$) may not satisfy $Q_{t+n} = q_m$, or in other words, the $p_i$ has an incorrect annotation. Training a model using data with wrong labels surely degrades the model performance. In the machine learning community, there have been many efforts in training a deep learning model with incorrectly labelled or unlabelled samples with weakly supervised learning approaches[12]–[16]. Such methods fit nicely into our situation.

We can define the training process formally as a weakly supervised learning problem. The task is to learn two functions:

$$\tau: \wp_{i,t} \to s_{i,t}; \theta \text{ and } \omega: [\tau(\wp_{i,t})] \to s_t; \hat{\theta}$$

($\theta$ and $\hat{\theta}$ denotes model parameters) from a training dataset:

$$\mathcal{D} = \{(I_{t_1}, s_{t_1}) ..., (I_{t_n}, s_{t_n})\},$$

$$\text{where } I_t = \{\wp_{1,t}, \wp_{2,t}, ..., \wp_{i,t}\}$$

---
[1] Unless otherwise indicated, in the rest of the paper, $I_t$ refers to the "virtual large image", which is a collection of patch-images taken by the camera.

$I_t$ is defined as a bag and $p_{i,t}$ are the instances from this bag. $I_t$ is a positive bag, i.e., $\omega(I_t) = s_t$, if the fusion of all its instances $p_{i,t}$ is positive ($\omega(\tau(p_{i,t})) = s_t$). The goal is to learn model parameters $\theta$ and $\hat{\theta}$ that maximise the data likelihoods:

$$\tau = \underset{\theta}{\mathrm{argmax}} \prod_{p_{i,t}, s_{i,t} \in \mathcal{D}} P(p_{i,t}, s_{i,t}|\theta) \quad (7.a)$$

$$\omega = \underset{\hat{\theta}}{\mathrm{argmax}} \prod_{p_{i,t}, s_t \in \mathcal{D}} P(\tau(p_{i,t}), s_t|\hat{\theta}) \quad (7.b)$$

To update $\theta$ and $\hat{\theta}$, we propose an algorithm that is based on the other two algorithms, namely, the expectation-maximisation (EM) algorithm[17] and the k-means clustering algorithm[18]. The k-means algorithm is applied first to find out the correct classes in stage one, followed by the EM algorithm for updating $\hat{\theta}$ values in stage two.

*The Algorithm: Stage One*

Human operators tend to use rough numbers for setting the feed load value, such as 125,130 and 145. Although these settings may satisfy the product quality constraint ($q_m$), there may exist better values which, although not being reflected in the training data, can maximise the equation (7. a) better. To verify this, we first apply a k-means algorithm to find the best possible clusters/classes.

1. Initial E Step: At the initial step, we initialise all the instance labels with the bag's label ($I_t, s_t$) from $\mathcal{D}$, and form the dataset :
$\mathcal{D}_1 = \{(p_{i,t}, s_t)_t, (p_{i,t_1}, s_{t_1})_{t_1}, \ldots, (p_{i,t_n}, s_{t_n})_{t_n}\}$ for all possible $I_t \in [I_{t_1}, I_{t_2}, I_{t_3}, \ldots, I_{t_n}]$
2. M Step: We then update the parameter $\theta$ to maximise equation (7. a) using the data from $\mathcal{D}_1$ till convergence.
3. E step: Through applying $\tau(p_i; \theta)$, we extract features $f_i$ for each $p_i$; store them with $p_i$'s original label and form a new dataset:
$\mathcal{D}_k = \{(f_{i,t}, s_{i,t})_t, (f_{i,t_1}, s_{i,t_1})_{t_1}, \ldots, (f_{i,t_n}, s_{1i})_{t_n}\}$
4. The k-means algorithm is then applied to $\mathcal{D}_k$ to form clusters with a range of initial k values.
5. We use the k-elbow method to select the best k value and perform k-means with this value to form clusters, $C_k$.
6. For each cluster, $c_i \in C_k$, we update the corresponding $(p_{i,t}, s_{i,t})$ in $\mathcal{D}_1$ with the mode value and generate the relabelled dataset :
$\mathcal{D}_1 = \{(p_{i,t}, \hat{s}_{i,t})_t, (p_{i,t_1}, \hat{s}_{i,t_1})_{t_1}, \ldots, (p_{i,t_n}, \hat{s}_{i,t_n})_{t_n}\}$.
7. We iterate back to step 2 till convergence. It should be noted, if new classes are generated in step 6, $\theta$ needs to be re-initialised.

The main goal for stage one is to regenerate labels for image patches $p_{i,t}$ automatically using their structural similarity and discover new labels that do not exist in the original dataset. In addition, a model is trained for making patch level predictions.

*The Algorithm: Stage Two*

In stage two, we start with a new CNN model instance, $\omega(s_t; \tau(p_{i,t}), \hat{\theta})$ that is trained using outputs from $\tau(p_{i,t})$.

8. We use the model, $\tau$, from stage one to transform all $p_{i,t}$ into $\tau(p_{i,t})$ and form a new dataset $\mathcal{D}_t$:
$\mathcal{D}_t = \{(\{\tau(p_{1,t}), \ldots, \tau(p_{i,t})\}, s_{t_1}), \ldots\}$
9. We update the parameter $\hat{\theta}$ to maximise equation (7. b) using the data from $\mathcal{D}_t$.

The second function $\omega$ fuses the representations of all instances ($p_{i,t}$) for a bag $I_t$. The representation format it takes depends on the outputs of the first model $\tau$. If the outputs of $\tau$ are target labels ($s_t$), the $\omega$ function can be as simple as a weighted average function or even a max-pooling function. The $\tau$ function, on the other hand, can produce more comprehensive representations of $p_{i,t}$. In this case, $\omega$ acts as a proper fusion model that combines all features derived from ($p_{i,t}$) and makes predictions.

### C. Model Architectures

*CNN Architecture for Patch Prediction (Function $\tau$)*

Unlike the other classification/segmentation tasks, ore pellet images do not contain straightforward features that a model can easily learn. For a task like this, significant amounts of training data are required for the model to pick up useful features. Although augmented data can be used, the number of final training samples still depends on the number of images collected before augmentation.

Another approach for increasing training data size, especially image data, is using image segments[2] [19]. This approach naturally fits our application. It is reasonable to assume that for a single ore pellet image patch, its segments have a similar distribution to the whole image. Also, training a model using overlapped segments has good augmentation and regularisation effects. A possible side-effect is that we may lose some global features at the image level. However, this can be overcome by combining feature maps from all segments to form global feature maps.

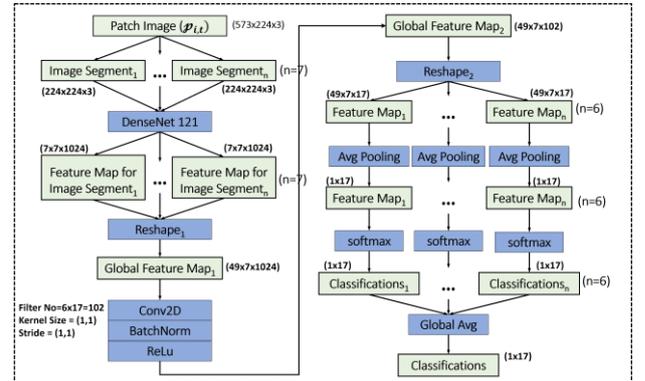

*Figure 4: Network Architecture of the Patch Prediction Model*

---

[2] We use the term "*segment*" here to distinguish ourselves from the "*patch*" term used in the previous section. It refers to the image patch concept in others' work.

Figure 4 shows the complete network architecture that we used for approximating function $\tau$. The boxes in lime are data passed into/out from each network layer, and the boxes in blue are network layers. The network's input is a set of patch image segments, and the network's output is one-hot-encoded classification results. The network has three main blocks: the encoder block, the feature aggregation block, and the multi-head classification block.

We use 121 layers DenseNet [20] block as the encoder for feature extraction for each input segment. We choose DenseNet over the other popular network architectures such as ResNet [21] because, with DenseNet, low-level features are pushed to the top of the network and the high-level ones. We find this property important for this work, as the size and colour of individual ore pellet may be key factors for making classification decisions. On the other hand, we still need filters with larger receptive fields that provide global views of the segment.

It is important to point out that the DenseNet block is applied to all the incoming segments simultaneously. There is only one DenseNet block instance in our network. During the training process, the weights of the DenseNet block get updated using all the gradients backpropagated from its next layer. Unlike that with a sequence model, no specific order is required for the input image segments. This can be understood as: the DenseNet block is trained N times in one backpropagation iteration with N data samples. Similarly, the same filters from the DenseNet block are applied to all segments for feature extraction and generate their feature maps separately during the feedforward phrase.

The feature maps of segments are aggregated into a global feature map (Global Feature Map$_1$), which then goes through a 2D convolutional (conv2d) layer with (1,1) kernel. The conv2d layer combines filter (1024) activations at different scales into the global ones for each receptive view (Global Feature Map$_2$). Afterwards, the final global feature map$_2$ is reshaped to multi-head feature maps, with each one corresponding to a classifier.

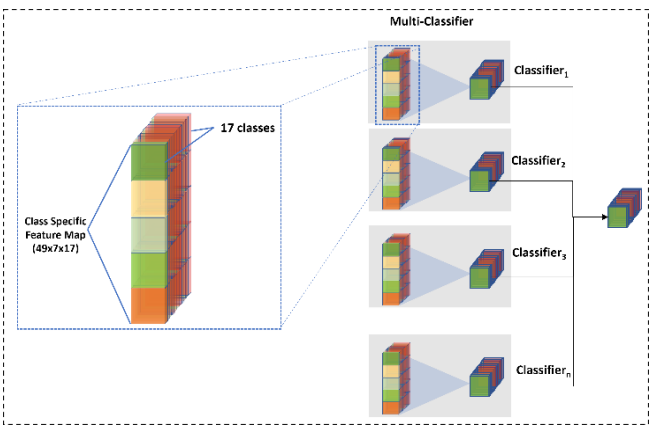

Figure 5: Enforcement of Class-Specific Feature Maps

Figure 5 gives an in-depth look at the multi-head classification block. There are two design rationales behind this. First, the features contributing to a particular class are aggregated together, which helps later model interpretability analysis through visualisation. Second, multi-head structure helps with the model generalisation and robustness. We observed that the multi-head classifier yields better classification results than a single classifier through our experiments.

*CNN Architecture for $I_t$ Prediction (Function $\omega$)*

The second network is for approximating the function $\omega$ (shown in Figure 6). It takes multiple global feature map$_2$ that are output by the first network and fuses them into one feature map for $I_t$. Due to the size of global feature map$_2$ and the number of patch images involved (must be big enough to cover the whole $I_t$ receptive field), we further reduce the feature map's size before they are combined into the final "$I_t$ feature map" through a convolutional layer.

There are two main reasons for using features maps from multiple patch images. First, although the first network predicts a class label for each patch image, the prediction results cannot be combined with mechanisms like voting or average pooling[22]–[24], as some class labels are generated from the clustering analysis and do not represent any real production setting.

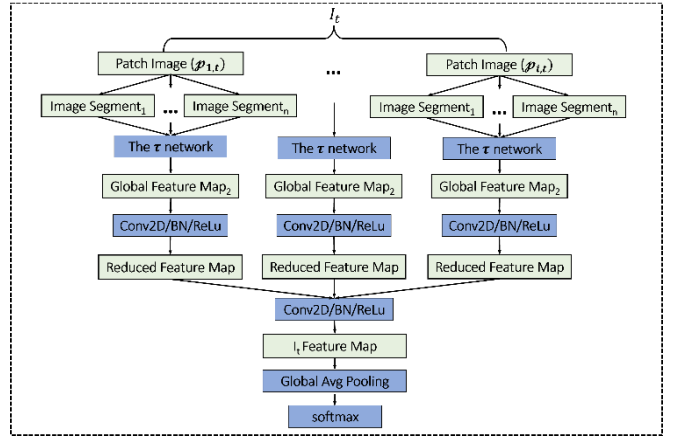

Figure 6: Network Architecture of the Image ($I_t$) Prediction Model

This is illustrated in more detail in the following sub-section. Also, patch level predications contain errors, which are brought to the final prediction if used directly. At last, the patch level model might be biased due to the smaller receptive field adopted, and the image level model may learn to correct the biased cases that occurred at the patch level.

## III. EXPERIMENTS

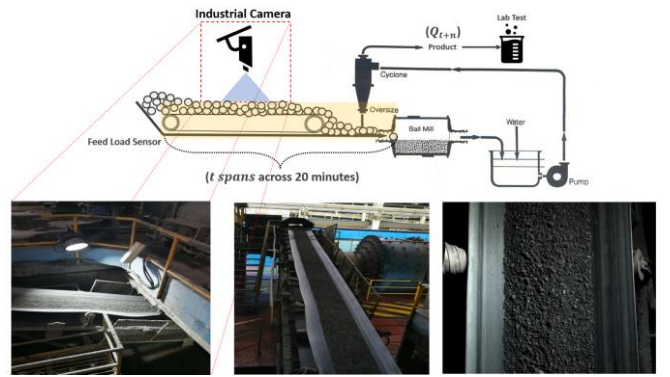

Figure 7: Experiment Setup for Data Collection

We evaluate our method through experiments in a real production environment. We use the Tensorflow[28]

framework for our model implementation, and the training of both models are performed on an Nvidia RTX 3090 GPU. Figure 7 shows the basic experimental setup.

*A. Data Collection and Pre-processing*

To train the two models, we collected three types of data in 14 months between June 2020 and August 2021. The first type of data is the lab test result ($Q_{t+n}$) of the final product quality, the second type is the feed load data ($s_t$) extracted from the equipment sensor readings, and the third type is the image ($p_{i,t}$) data taken by an industrial camera mounted perpendicularly to the conveyor belt.

All three types of data are timestamped. Based on the minimum product quality requirements, the lab test results that are below 66 are dropped. It ensures the condition $Q_{t+n} = q_m = 66$ in equation (6). For each filtered test result Q at time t+n, we find its corresponding feed load data at time t with the value of n set to 90 minutes. It is worth mentioning that the data stamped at time t is not a single sample reading but an average reading of a collection of samples. The sampling rate of the feed load sensor is one minute, but it takes around 20 to 30 minutes for the conveyor to transmit ore pellets for producing the final product at the time $Q_{t+n}$. Therefore, we aggregated all feed load sensor readings within that 20-minutes window into a single reading $s_t$. The same principle applies to the image data. Our camera captures ore pellet images (1024x400x3) with the same sampling rate as the feed load sensor. All images ($p_{i,t}$) in the same 20-minutes window are grouped to form the virtual image ($I_t$), and are all annotated with a label $s_t$ initially.

In total, we collected 29,952 samples (($p_{i,t}, s_t$)) as the dataset $D_1$ for training/validating the first model. From the datasets, we extracted 17 classes (0, 120, 125, 130, 132,135, 136, 137, 138, 139, 140, 141, 142, 143, 144, 145, 146) using unique $s_t$ values. The class '0' indicates no ore pellets on the conveyor, while each of the rest represents a possible feed load setting value. The patch images are all downsized to 573x224x3. Seven segments of size 224x224x3 are then extracted from each patch image with a stride value of 56 pixels. For the data augmentation purposes, segments are randomly flipped either horizontally or vertically, and the brightness values of the segments are also randomly adjusted between 0.5 and 1. 80% of data are used for training, with the remaining 20% for testing.

*B. Training the First Model*

Following the proposed algorithm, we start with training the model using the initial dataset $D_1$. The Stochastic Gradient Descent (SGD)[29] is used as our optimiser with an initial learning rate of $10^{-2}$ and the momentum value of 0.9. The learning rate is gradually reduced to $10^{-4}$ with a deduction rate of 0.7, given that there has been no improvement in the past five epochs. The batch size used for training is 32. The training process converged around epoch 80, with the best validation accuracy of 87.82%. We then used this provisional model for extracting features from all patch images in $D_1$ and performed the principal component analysis[30] for feature reduction. We choose to keep the first 250 components as they explain 95% variance of the original feature space (Figure 8. a).

A series of k-means clustering analysis were applied afterwards with the value of k∈ [10,31]. The k-elbow method shows the best k value is 18 (Figure 8. b), which indicates there exist one additional possible class on top of the 17 initial classes from our dataset.

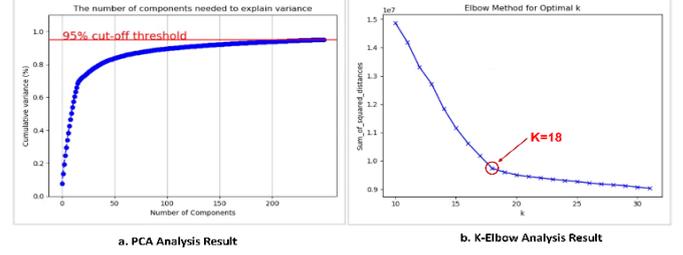

*Figure 8: The PCA and K-Elbow Analysis Results*

Figure 9 shows that for the first 17 clusters (C1-C17), most of their members have consistent class labels corresponding to the original ones. However, for cluster C18, the class labels of its members are diverse. Such facts suggest that the patch images inside the C18 cluster have significant structural differences with the images from the other clusters; hence the initial class labels assigned to them may be inaccurate. Therefore, we reconstructed another dataset $D_{1u}$ via regrouping the patch images based on their clusters and used the cluster labels as their class labels. $D_{1u} = \{(p_{1,t}, C_1), …, (p_{i,t}, C_n)\}$.

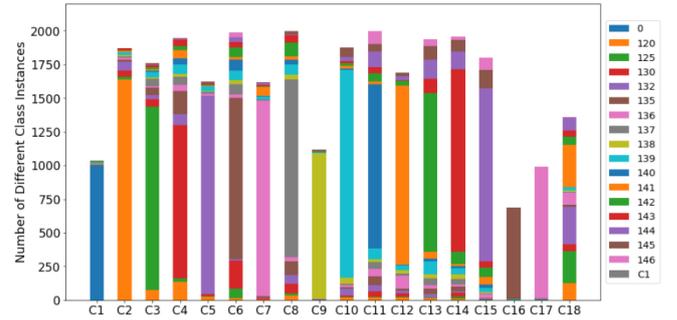

*Figure 9: Number of Class Instances Within Each Cluster*

We trained and validated a new network model using the $D_{1u}$ dataset (with 80%/20% split) in four iterations. This model is like the first one, apart from being used for an 18-classes classification task. After each iteration, we relabelled the patch images using the predicated class labels and updated the dataset $D_{1u}$ for the next iteration. The training accuracy for every class reached 100% after the fourth iteration, and the overall validation accuracy of the model increased from 88.12% (after the first iteration) to 96.45% (after the fourth iteration). We tried to conduct clustering analysis between each iteration using the latest model, but the number of generated clusters remained 18, and the cluster members did not vary significantly.

*C. Training the Second Model*

For the second model, we grouped patch images ($p_{(i,t)}$) from the dataset $D_1$ into ($I_t, s_t$) pairs for five different datasets $D_2 = \{D_{2,15mins}, D_{2,20mins}, D_{2,25mins}, D_{2,30mins}, D_{2,35mins}\}$. In practice, only a rough estimation is instructed on how long a ball mill is filled from empty to full. The ball mill is filled quicker for higher feed-load settings, while it takes longer for the lower feed-load settings. Therefore, we must try different window sizes to determine the best size for $I_t$. It also should be noted that using datasets with different input sizes has little impact on the first model, as such differences only cause small vibrations in the sample size of $D_1$. It is reasonable to assume that the performance of the first model at the patch image level remains the same for different $D_2$s. We trained five instances of the second model, each using a different $D_2$ dataset. The same hyper-parameter values for training the first model are used for training all five instances.

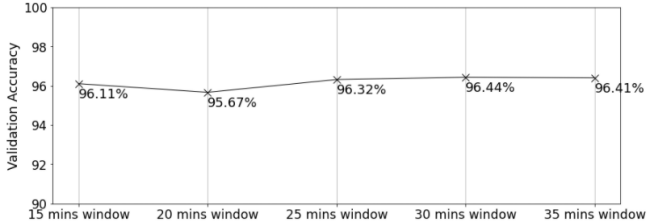

*Figure 10: Test Accuracies for Model Instances Trained with Five Different Window Sizes*

Figure 10 shows no significant difference between the instances' performance, with the best test accuracy (96.44%) achieved for the 30-minutes window and the worst (95.67%) for the 20-minutes window.

| Classes | Precision | Recall | F1 | Most Misclassified |
|---------|-----------|--------|------|--------------------|
| 0       | 1.00      | 1.00   | 1.00 | -                  |
| 120     | 0.96      | 0.99   | 0.98 | 143: 0.1%          |
| 125     | 1.00      | 0.98   | 0.99 | 137: 0.67%         |
| 130     | 0.95      | 0.98   | 0.96 | 135: 0.54%         |
| 132     | 0.97      | 0.97   | 0.97 | 120: 1.2%          |
| 135     | 0.95      | 0.97   | 0.96 | 130: 1.56%         |
| 136     | 0.96      | 0.96   | 0.96 | 139: 0.56%         |
| 137     | 0.96      | 0.96   | 0.96 | 139: 0.61%         |
| 138     | 0.97      | 0.97   | 0.97 | 139: 2.31%         |
| 139     | 0.95      | 0.94   | 0.94 | 138: 1.06%         |
| 140     | 0.91      | 0.98   | 0.95 | 130: 0.63%         |
| 141     | 0.92      | 0.98   | 0.95 | 136: 1.15%         |
| 142     | 0.96      | 0.94   | 0.95 | 140: 1.74%         |
| 143     | 0.96      | 0.97   | 0.96 | 135: 0.69%         |
| 144     | 0.89      | 0.96   | 0.92 | 141: 0.82%         |
| 145     | 0.98      | 0.79   | 0.87 | 144: 7%, 141: 6.7% |
| 146     | 1.00      | 0.89   | 0.94 | 144: 6.75%         |

*Table 2: Classification Results of the Best Model*

We also used the precision, recall, and F1 score to measure each class's model's performance. The results (only for the best model due to the paper space limits) are shown in Table 2. It also shows the proportion of most misclassified class instances against each ground-truth class, from which we can see that the worst two recall values are from the classes 145 (79%) and 146 (0.89%). 13.7% of the testing 145 class are misclassified as class 144 (7%) and class 141 (6.7%), while 6.75% of class 146 are misclassified as class 144.

There are three possible reasons for the descending performances of these three classes (144,145,146). First, the dataset may be unbalanced, which is not our case. Second, the features of these three classes are too close to be differentiated. If this is the case, further refinements (model architecture change, parameter turning) may be necessary for better feature extractions. However, as the precision values of these classes are high, which indicates the number of false positive cases is low, we have reasons to believe that the features used for classification are well learnt.

Another possible reason is the wrong data labelling, which is more likely to be our case. In the production environment for our work, it is not an unusual case that a human operator leaves the feed load unchanged even if the final product quality is a lot better than the quality threshold ($Q_{t+n} > 66$). For such circumstances, the data labelling (($\mathcal{P}_{i,t}, s_t$)) is incorrect, as there should exist a larger feed load ($s'_t > s_t$) that can be used as the correct label for $\mathcal{P}_{i,t}$. Unfortunately, it is not possible to find the value of $s'_t$ as it is never set in the system.

### D. Model Interpretability and Visualisation

Even though the model shows promising classification results with the testing datasets, it is still essential to understand what the model has learnt and how it makes its decision. To visualise what the model has learnt, we adopt the activation maximisation method [31]. We first visualise the "global feature map$_2$" from the first model, as it aggregates all features learnt before they get enforced into class-specific feature maps.

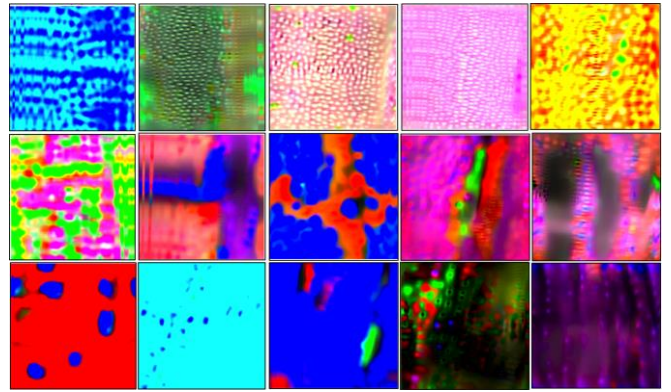

*Figure 11: Visualisation of Samples in the Global Feature Map$_2$*

From sampled feature maps shown in Figure 11, we can see that, to make a prediction, the model considers several factors at both ore pellet level and patch image level. The top row shows that the model focuses on the ore pellets level regarding their colour, sizes, and overall quantities, whereas the feature maps in the second row focus more on global shapes at the patch image level. Those shown in the third row are mainly colour channels. Since the following layers combine all these different features from the network for making predictions, it is reasonable to conclude that these are the key factors that affect the final classification result. It is consistent with the knowledge from experienced mineralogists who can tell ore's characteristics via looking at their sizes and colours

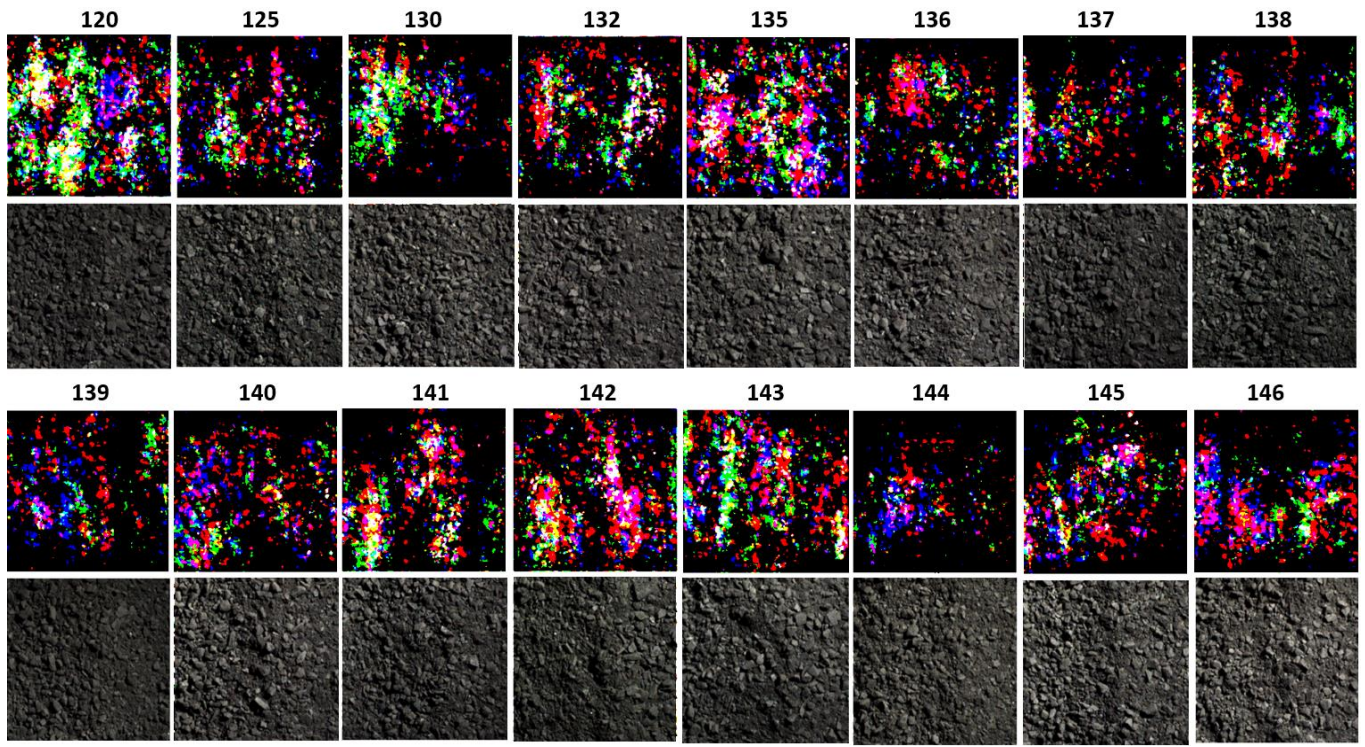

*Figure 12: Saliency Maps for the Best True Positive Class Samples. The top image is the saliency map for each class, and the bottom is the ore pellet image. All the saliency map images are adjusted for their sharpness to give better visibility*

To give a more intuitive impression of what the model looks at for the final classification, we have also used the saliency map[32] to outline the model's attention areas (shown in Figure 12). For each class, we selected the best true positive sample (the image with the correct and the highest classification score) for maximising the saliency map. The image channels (RGB) are kept for this visualisation task. For all saliency maps, image areas/channels with significant contributions to the classification are highlighted. The different colours in the maps potentially highlight different types of ores (hematite: red/purple, chlorite: green, limonite: brown and magnetite: dark). As hematite and limonite are easier to grind and contain higher amounts of iron, the feed load can be set higher for ore pellets mainly of these two types.

the same product quality as the other two ore types. The generated saliency maps are aligned with such facts. For ore pellet images with lower feed load settings (from 120 to 132), a good proportion of highlighted areas are coloured green, while for middle and high range feed load settings (from 135 to 146), red/purple/pink colours become dominant.

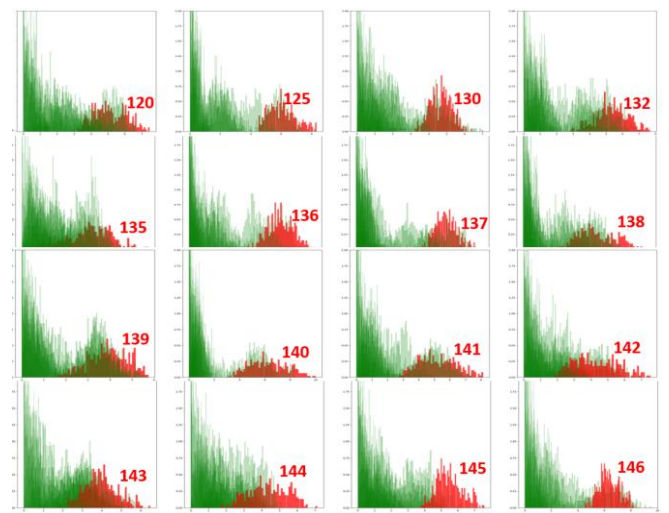

*Figure 14: Distributions of the Enforced Class-Specific Feature Maps Activated by the Worst True Positive Samples*

At last, we show how the model makes classification decisions based on the learnt features. We plot the distribution of its "enforced class-specific feature map" for each class, using the best and the worst true positive samples, respectively. For the best cases (Figure 13), there are clear decision boundaries between the feature distributions of the predicated class and the other classes, whereas, for the worst

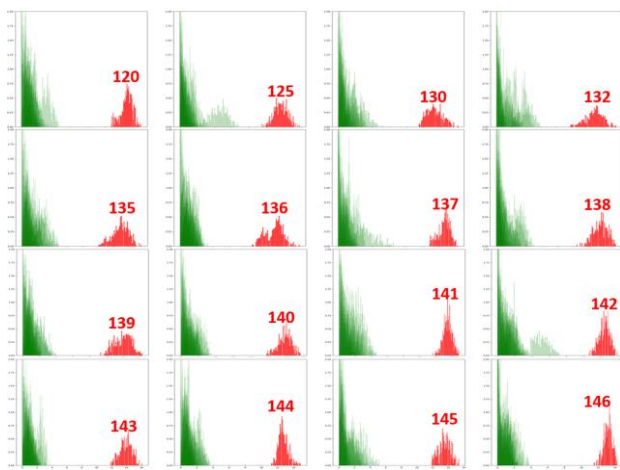

*Figure 13: Distributions of the Enforced Class-Specific Feature Maps Activated by the Best True Positive Samples*

In contrast, for chlorite (containing less iron) and magnetite (hard to grind), the feed load needs to be set lower to ensure cases (Figure 14), such decision boundaries are less determinate.

IV. EVALUATION

The difficulties in evaluating this work stem from several facts. First, there is no standard bench-test dataset available. Image datasets from the literature [5], [33] are taken of ore rocks before crushing. Comparing our work with those does not lead to meaningful conclusions. Second, to the best of our knowledge, our work is the first of its kind that estimates feed load setting for the optimisation purpose. Most of the existing work [3], [5], [6], [33]–[37] are focused on classification or segmentation of ore grade, ore shape and ore characteristic, hence are neither directly comparable. Third, it is infeasible to use our model in a separate production line and compare its performance with that from a human-operated environment due to the risk of potential economic loss if the model fails to work. This restriction makes it hard to carry statistical tests. Finally, in the actual production environment, the feed load adjusted by human operators are often less optimal. Comparing the model prediction results with actual feed load settings may lead to confusion. For instance, a feed load setting of 125 tons may produce a satisfiable product quality ($Q_{t+n} \geq 66$). A higher feed load setting may also meet this condition but is never set by human operators. In this case, our model may give a correct optimal estimation with a higher feed load setting but is different from the ground truth label.

Due to the above issues and restrictions, we have to use a compromised evaluation strategy. Between 7$^{th}$ and 30$^{th}$ September 2021, we ran our model in parallel with the existing grinding production system and collected 23 days of data for evaluation. For every lab test result, we compare the actual feed load and our model prediction (both are from 2 hours earlier before the lab test time). For lab test results that meet the minimal product quality requirement ($\geq 66$), if our model prediction value is equal or greater than the true feed load setting, we consider it as a correct ($MP_c$) or possible correct prediction ($MP_c+$), as if the actual feed load was adjusted higher, we may still achieve the same product quality but with better productivity. For lab test results that miss the minimal quality requirement, if our model prediction value is smaller than the true feed load setting, it is also considered as a possible correct prediction ($MP_c-$) as lowering the feed load may lead to better product quality. All other cases are considered as wrong predictions ($MP_w$).

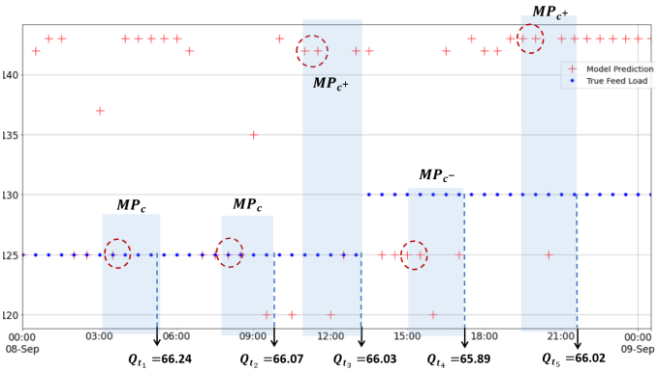

*Figure 15: Comparison Result between the Model Prediction and the True Feed Load Setting*

Figure 15 shows the comparison results between our model and the actual feed load setting for a single working day. For the first two lab test results ($Q_{t_1} > 66, Q_{t_2} > 66$), our model agreed with the human operator's feed load setting at 125. For the third lab test results ($Q_{t_3} > 66$), the model believed the feed load should be set to 142, while the human operator left it to 125. Interestingly, the human operator lifted the feed load to 130 after the third lab test result was generated, as he probably noticed that all the last three $Q_{ts}$ are satisfactory hence trying to increase the throughput. Unfortunately, such a change led to poor product quality, as reflected by the lab test results $Q_{t_4} < 66$. The model, however, suggested setting the feed load lower at 125 for that period. For $Q_{t_5} > 66$, there could be a throughput increment based on the model's output (set feed load to 144), but the actual operation just left everything untouched.

In summary, for the whole evaluation period (24 days), we collected 144 lab test results, with 89 of them greater than 66 and the rest 55 are less. It indicates that the human operators were only able to keep the product quality qualified in 61.8% of the time, not talking about whether the throughputs for these qualified products are optimal. In total, human operators made 67 changes to the feed load setting, which is on average 2 to 3 times a day. In contrast, our model suggested 577 feed load changes (if two consecutive predictions have the same value, it is not counted as a change) that is averagely 24 times a day or once an hour. The $MP_C, MP_C+, MP_C-$ and $MC_w$ values are 23.2%, 43.7%, 29.1% and 4% respectively. The total throughputs of qualified products are 11,419 tons based on humans' operations, and 12,827 tons should follow the model's predictions. It yields 12.3% throughput boosting for qualified products, which potentially has a huge economic impact on the organisation.

V. CONCLUSIONS

This paper presented an image-based solution for direct feed load estimation in grinding production in the mineral industry. Deep learning models and weakly supervised learning methods are the primary tools used. We proposed to use a two-stage training algorithm that first utilises inaccurately labelled patch images for feature extraction and then fuses extracted features for optimal feed load estimation in real-time. Through experiments, we showed a detailed demonstration of how the proposed algorithm works and provided comprehensive visualisation and discussion, showing what the model has learned and aligned with mineralogists' knowledge. We also discussed our evaluation approach and demonstrated the performance gain with the proposed work, comparing to human operators. Putting it all together, we are confident that this proposed work makes the whole grinding process much more productive and potentially brings substantial economic benefits to the mineral organisations.

We will seek methods that could generalise and transfer a learnt model to other production lines in the future. Due to the geographical distribution, the ore characteristics at different production lines vary significantly. A model trained on one production line is unlikely to scale. It requires large amounts of time for data collection and repetitive model training for such circumstances, which may become a barrier for the broader deployment of this work. We also plan to investigate the lightweight model architectures of the first model. Although the first model is used for feature extractions in this work, it outputs feed load prediction for a single patch image. The result will help provide instant feedback in production if the model can be deployed on portable devices in the environment without a network connection. It is not practical due to the current size and complexity of the model at present. Therefore, a mobile-friendly network architecture that does not compromise with too much performance

References

[1] L. Tuşa, M. Kern, M. Khodadadzadeh, R. Blannin, R. Gloaguen, and J. Gutzmer, "Evaluating the performance of hyperspectral short-wave infrared sensors for the pre-sorting of complex ores using machine learning methods," *Minerals Engineering*, vol. 146, p. 106150, Jan. 2020, doi: 10.1016/J.MINENG.2019.106150.

[2] C. Robben, P. Condori, A. Pinto, R. Machaca, and A. Takala, "X-ray-transmission based ore sorting at the San Rafael tin mine," *Minerals Engineering*, vol. 145, p. 105870, Jan. 2020, doi: 10.1016/J.MINENG.2019.105870.

[3] G. Li, B. Klein, C. Sun, and J. Kou, "Applying Receiver-Operating-Characteristic (ROC) to bulk ore sorting using XRF," *Minerals Engineering*, vol. 146, p. 106117, Jan. 2020, doi: 10.1016/J.MINENG.2019.106117.

[4] "Deep Learning - Ian Goodfellow, Yoshua Bengio, Aaron Courville - Google Books." https://books.google.co.uk/books?hl=en&lr=&id=omivDQAAQBAJ&oi=fnd&pg=PR5&dq=deep+learning&ots=MNP0iorDVS&sig=f15TevuKDsA2j9NK3u9HZHZT_lA&redir_esc=y#v=onepage&q=deep%20learning&f=false (accessed Sep. 28, 2021).

[5] V. Singh and S. Mohan Rao, "Application of image processing and radial basis neural network techniques for ore sorting and ore classification," *Minerals Engineering*, vol. 18, no. 15, pp. 1412–1420, Dec. 2005, doi: 10.1016/J.MINENG.2005.03.003.

[6] Y. Liu, Z. Zhang, X. Liu, L. Wang, and X. Xia, "Deep learning-based image classification for online multi-coal and multi-class sorting," *Computers & Geosciences*, vol. 157, p. 104922, Dec. 2021, doi: 10.1016/J.CAGEO.2021.104922.

[7] E. Gülcan and Ö. Y. Gülsoy, "Performance evaluation of optical sorting in mineral processing – A case study with quartz, magnesite, hematite, lignite, copper and gold ores," *International Journal of Mineral Processing*, vol. 169, pp. 129–141, Dec. 2017, doi: 10.1016/J.MINPRO.2017.11.007.

[8] C. A. Perez *et al.*, "Ore grade estimation by feature selection and voting using boundary detection in digital image analysis," 2011, doi: 10.1016/j.minpro.2011.07.008.

[9] J. Maitre, K. Bouchard, and L. P. Bédard, "Mineral grains recognition using computer vision and machine learning," *Computers & Geosciences*, vol. 130, pp. 84–93, Sep. 2019, doi: 10.1016/J.CAGEO.2019.05.009.

[10] E. Donskoi *et al.*, "Utilisation of optical image analysis and automatic texture classification for iron ore particle characterisation," 2007, doi: 10.1016/j.mineng.2006.12.005.

[11] Y. Liu, Z. Zhang, X. Liu, W. Lei, and X. Xia, "Deep Learning Based Mineral Image Classification Combined with Visual Attention Mechanism," *IEEE Access*, vol. 9, pp. 98091–98109, 2021, doi: 10.1109/ACCESS.2021.3095368.

[12] K. Dong and D. Jiang, "Automated Estimation of Ore Size Distributions Based on Machine Vision," *Lecture Notes in Electrical Engineering*, vol. 238 LNEE, pp. 1125–1131, 2014, doi: 10.1007/978-1-4614-4981-2_122.

[13] E. Hamzeloo, M. Massinaei, and N. Mehrshad, "Estimation of particle size distribution on an industrial conveyor belt using image analysis and neural networks," *Powder Technology*, vol. 261, pp. 185–190, Jul. 2014, doi: 10.1016/J.POWTEC.2014.04.038.

[14] E. Donskoi, T. D. Raynlyn, and A. Poliakov, "Image analysis estimation of iron ore particle segregation in epoxy blocks," *Minerals Engineering*, vol. 120, pp. 102–109, May 2018, doi: 10.1016/J.MINENG.2018.02.024.

[15] Y. Liu, Z. Zhang, X. Liu, L. Wang, and X. Xia, "Performance evaluation of a deep learning based wet coal image classification," *Minerals Engineering*, vol. 171, p. 107126, Sep. 2021, doi: 10.1016/J.MINENG.2021.107126.

[16] J. A. Herbst, W. T. Pate, and A. E. Oblad, "Model-based control of mineral processing operations," *Powder Technology*, vol. 69, no. 1, pp. 21–32, Jan. 1992, doi: 10.1016/0032-5910(92)85004-F.

[17] X. Chen, J. Zhai, Q. Li, and S. Fei, "Fuzzy Logic Based Online Efficiency Optimisation Control of a Ball Mill Grinding Circuit," in *Fourth International Conference on Fuzzy Systems and Knowledge Discovery (FSKD 2007)*, 2007, pp. 575–580. doi: 10.1109/FSKD.2007.329.

[18] J. B. Yianatos, M. A. Lisboa, and D. R. Baeza, "Grinding capacity enhancement by solid concentration control of hydrocyclone underflow," *Minerals Engineering*, vol. 15, no. 5, pp. 317–323, May 2002, doi: 10.1016/S0892-6875(02)00027-4.

[19] L. Guo, H. Wang, and J. Zhang, "Data-driven grinding control using reinforcement learning," *Proceedings - 21st IEEE International Conference on High Performance Computing and Communications, 17th IEEE International Conference on Smart City and 5th IEEE International Conference on Data Science and Systems, HPCC/SmartCity/DSS 2019*, pp. 2817–2824, Aug. 2019, doi: 10.1109/HPCC/SMARTCITY/DSS.2019.00395.

[20] X. Lu, B. Kiumarsi, T. Chai, F. L.-I. C. T. & Applications, and undefined 2016, "Data-driven optimal control of operational indices for a class of industrial processes," *IET*.

[21] S. Yin, H. Gao, and O. Kaynak, "Data-Driven Control and Process Monitoring for Industrial Applications—Part I," *IEEE Transactions on Industrial Electronics*, vol. 61, no. 11, pp. 6356–6359, Nov. 2014, doi: 10.1109/TIE.2014.2312885.

[22] Y. Lecun, Y. Bengio, and R. 4g332, "Convolutional Networks for Images, Speech, and Time-Series".

[23] Z.-H. Zhou, "A brief introduction to weakly supervised learning," *National Science Review*, vol. 5, no. 1, pp. 44–53, Jan. 2018, doi: 10.1093/NSR/NWX106.

[24] Y. F. Li, L. Z. Guo, and Z. H. Zhou, "Towards Safe Weakly Supervised Learning," *IEEE Transactions on Pattern*

*Analysis and Machine Intelligence*, vol. 43, no. 1, pp. 334–346, Jan. 2021, doi: 10.1109/TPAMI.2019.2922396.

[25] M. Oquab, L. Bottou, I. Laptev, and J. Sivic, "Is object localisation for free? – Weakly-supervised learning with convolutional neural networks," Jun. 2015, Accessed: Sep. 17, 2021. [Online]. Available: https://hal.inria.fr/hal-01015140

[26] J. E. van Engelen and H. H. Hoos, "A survey on semi-supervised learning," *Machine Learning 2019 109:2*, vol. 109, no. 2, pp. 373–440, Nov. 2019, doi: 10.1007/S10994-019-05855-6.

[27] X. Zhai, A. Oliver, A. Kolesnikov, and L. Beyer, "S4L: Self-Supervised Semi-Supervised Learning." pp. 1476–1485, 2019.

[28] T. K. Moon, "The expectation-maximisation algorithm," *IEEE Signal Processing Magazine*, vol. 13, no. 6, pp. 47–60, 1996, doi: 10.1109/79.543975.

[29] J. A. Hartigan and M. A. Wong, "Algorithm AS 136: A K-Means Clustering Algorithm," *Applied Statistics*, vol. 28, no. 1, p. 100, 1979, doi: 10.2307/2346830.

[30] T. Oga, R. Harakawa, S. Minewaki, Y. Umeki, Y. Matsuda, and M. Iwahashi, "River state classification combining patch-based processing and CNN," *PLOS ONE*, vol. 15, no. 12, p. e0243073, Dec. 2020, doi: 10.1371/JOURNAL.PONE.0243073.

[31] G. Huang, Z. Liu, L. van der Maaten, and K. Q. Weinberger, "Densely Connected Convolutional Networks," *Proceedings - 30th IEEE Conference on Computer Vision and Pattern Recognition, CVPR 2017*, vol. 2017-January, pp. 2261–2269, Aug. 2016, Accessed: Sep. 19, 2021. [Online]. Available: https://arxiv.org/abs/1608.06993v5

[32] A. Cruz-Roa *et al.*, "Automatic detection of invasive ductal carcinoma in whole slide images with Convolutional Neural Networks," *Medical Imaging*, vol. 9041, pp. 904103–904104, 2014, doi: 10.1117/12.2043872.

[33] A. Seff *et al.*, "2D View Aggregation for Lymph Node Detection Using a Shallow Hierarchy of Linear Classifiers," *Medical image computing and computer-assisted intervention : MICCAI ... International Conference on Medical Image Computing and Computer-Assisted Intervention*, vol. 17, no. 0 1, p. 544, 2014, Accessed: Sep. 22, 2021. [Online]. Available: /pmc/articles/PMC4350911/

[34] S. Poria, E. Cambria, and A. Gelbukh, "Deep Convolutional Neural Network Textual Features and Multiple Kernel Learning for Utterance-level Multimodal Sentiment Analysis," *Conference Proceedings - EMNLP 2015: Conference on Empirical Methods in Natural Language Processing*, pp. 2539–2544, 2015, doi: 10.18653/V1/D15-1303.

[35] M. Abadi *et al.*, *TensorFlow: A System for Large-Scale Machine Learning*, vol. 10, no. July. 2016. Accessed: Sep. 22, 2021. [Online]. Available: https://www.usenix.org/system/files/conference/osdi16/osdi16-liu.pdf%5Cnhttps://www.usenix.org/conference/osdi16/technical-sessions/presentation/abadi

[36] L. Bottou, "Large-Scale Machine Learning with Stochastic Gradient Descent," *Proceedings of COMPSTAT 2010 - 19th International Conference on Computational Statistics, Keynote, Invited and Contributed Papers*, pp. 177–186, 2010, doi: 10.1007/978-3-7908-2604-3_16.

[37] S. Wold, K. Esbensen, and P. Geladi, "Principal component analysis," *Chemometrics and Intelligent Laboratory Systems*, vol. 2, no. 1–3, pp. 37–52, Aug. 1987, doi: 10.1016/0169-7439(87)80084-9.

[38] Y. Bengio, A. Courville, D. Erhan, Y. Bengio, and P. Vincent, "Visualizing Higher-Layer Features of a Deep Network Visualizing Higher-Layer Features of a Deep Network Département d'Informatique et Recherche Opérationnelle," 2009, Accessed: Sep. 26, 2021. [Online]. Available: https://www.researchgate.net/publication/265022827

[39] K. Simonyan, A. Vedaldi, and A. Zisserman, "Deep Inside Convolutional Networks: Visualising Image Classification Models and Saliency Maps," *2nd International Conference on Learning Representations, ICLR 2014 - Workshop Track Proceedings*, Dec. 2013, Accessed: Sep. 27, 2021. [Online]. Available: https://arxiv.org/abs/1312.6034v2

[40] J. Tessier, C. Duchesne, and G. Bartolacci, "A machine vision approach to online estimation of run-of-mine ore composition on conveyor belts," *Minerals Engineering*, vol. 20, no. 12, pp. 1129–1144, Oct. 2007, doi: 10.1016/J.MINENG.2007.04.009.

[41] D. P. Mukherjee, Y. Potapovich, I. Levner, and H. Zhang, "Ore image segmentation by learning image and shape features," *Pattern Recognition Letters*, vol. 30, no. 6, pp. 615–622, Apr. 2009, doi: 10.1016/J.PATREC.2008.12.015.

[42] C. A. Perez *et al.*, "Ore grade estimation by feature selection and voting using boundary detection in digital image analysis," *International Journal of Mineral Processing*, vol. 101, no. 1–4, pp. 28–36, Nov. 2011, doi: 10.1016/J.MINPRO.2011.07.008.

[43] N. Mustafa, J. Zhao, Z. Liu, Z. Zhang, and W. Yu, "Iron ORE Region Segmentation Using High-Resolution Remote Sensing Images Based on Res-U-Net," *International Geoscience and Remote Sensing Symposium (IGARSS)*, pp. 2563–2566, Sep. 2020, doi: 10.1109/IGARSS39084.2020.9324218.